%% file: main.tex
\documentclass[10pt,twocolumn,letterpaper]{article}

\PassOptionsToPackage{table}{xcolor}

\input{preamble}

\title{Constella: Learning What Goes Where through Configurational Binding in Vision-Language Models}

\author{Ziyu Fan\textsuperscript{1,*}\quad
  Junming Liu\textsuperscript{2,*}\quad
  Dingshuo Chen\textsuperscript{1}\quad
  Liuji Chen\textsuperscript{1}\quad
  Junfei Wu\textsuperscript{1}\\[3pt]
  Ximing Bai\textsuperscript{3}\quad
  Yufu Chen\textsuperscript{4}\quad
  Yiming Cheng\textsuperscript{5}\quad
  Weiping Ding\textsuperscript{6}\quad
  Shu Wu\textsuperscript{1,\textdagger}\\[5pt]
  {\small
    \textsuperscript{1}Institute of Automation, Chinese Academy of Sciences\quad
    \textsuperscript{2}Tongji University}\\
  {\small
    \textsuperscript{3}University of Chinese Academy of Sciences\quad
    \textsuperscript{4}University of California, Berkeley}\\
  {\small
    \textsuperscript{5}Tsinghua University\quad
    \textsuperscript{6}Nantong University}\\[4pt]
  {\small \textsuperscript{*}These authors contributed equally and share first authorship.\qquad Email: \href{mailto:fionaf111@163.com}{\texttt{fionaf111@163.com}}}\\
  {\small \textsuperscript{\textdagger}Corresponding author}}

\makeatletter
\let\constella@maketitle\@maketitle
\renewcommand{\@maketitle}{\constella@maketitle
  \noindent\begin{minipage}{\textwidth}
    \centering
    \includegraphics[width=\textwidth]{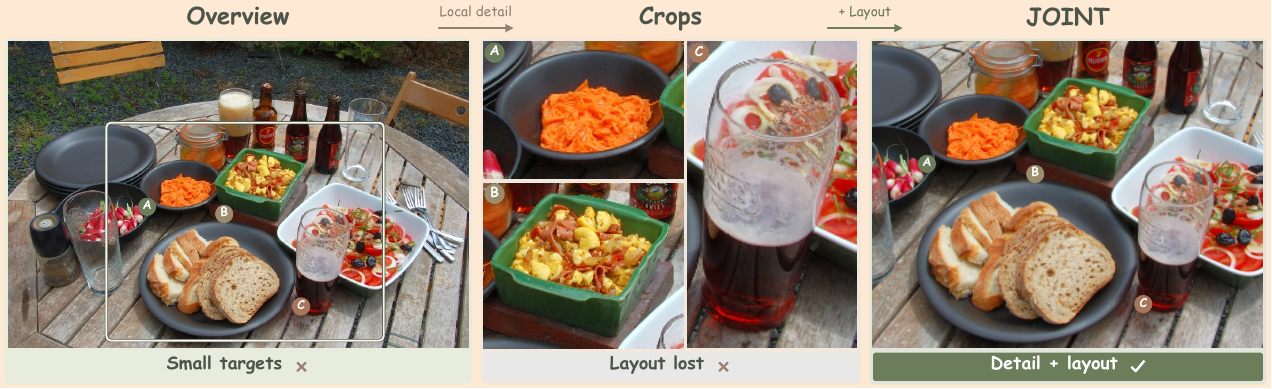}

    \captionsetup{hypcap=false}
    \captionof{figure}{\textbf{Preserving object details and shared spatial layout.}
        The overview retains the scene layout but can obscure small
        object details. Individual crops reveal these details without
        a shared spatial frame. JOINT enlarges the referred objects
        while preserving their relative layout.
        These privileged views are used during training.
        At inference, the student receives only the original image
        and question.
    }
    \label{fig:joint-teaser}
  \end{minipage}\par
}
\makeatother

\begin{document}

\maketitle

\input{0_abstract}
\input{1_intro}
\input{2_related_work.tex}
\input{3_methods}
\input{4_experiments}
\input{5_conclusion}

{\small

\input{main.bbl}
}

\onecolumn
\appendix
\input{supplement.tex}

\end{document}

%% file: preamble.tex
\ifdefined\pdfminorversion
\fi
\usepackage{constella_layout}
\ifdefined\XeTeXversion
  \usepackage{fontspec}
\fi
\usepackage{graphicx}
\usepackage{cuted}

\definecolor{bindQwen}{HTML}{E4EDF8}
\definecolor{bindOther}{HTML}{E8F3EA}
\definecolor{bindSpatial}{HTML}{F0EAF7}
\definecolor{bindOurs}{HTML}{FBE9DD}

\definecolor{constellalinkblue}{rgb}{0.21,0.49,0.74}
\usepackage[breaklinks,colorlinks,allcolors=constellalinkblue]{hyperref}

 \par
\usepackage{cuted}
\usepackage{currfile}
\usepackage{caption}

\usepackage{graphicx}
\usepackage{amsmath,amssymb}
\usepackage{booktabs,array,tabularx,textcomp}

\usepackage{colortbl}

\usepackage{float}
\usepackage{enumitem}

\makeatletter
\@ifundefined{algorithm}{\floatstyle{ruled}\newfloat{algorithm}{tbp}{loa}\floatname{algorithm}{Algorithm}}{}

\@ifundefined{algorithmsteps}{\newlist{algorithmsteps}{enumerate}{1}}{}
\makeatother

\setlist[algorithmsteps]{
    label=\arabic*:,
    ref=\arabic*,
    leftmargin=2em,
    labelsep=0.5em,
    itemsep=2pt,
    parsep=0pt,
    topsep=3pt
}

\definecolor{backboneband}{RGB}{237,242,247}
\definecolor{otherband}{RGB}{243,244,246}
\definecolor{spatialband}{RGB}{237,243,241}
\definecolor{oursband}{RGB}{232,239,248}

\usepackage{multicol}
\input{supplement/case_layout}

%% file: supplement/case_layout.tex
\usepackage{tcolorbox}
\definecolor{caseblue}{HTML}{EAF1F6}
\definecolor{caseblueborder}{HTML}{C4D5E1}
\definecolor{casegreen}{HTML}{EAF2ED}
\definecolor{casegreenborder}{HTML}{C6DACF}
\definecolor{caseheading}{HTML}{1B4259}
\definecolor{caseink}{HTML}{172C36}
\definecolor{wrongbg}{HTML}{F4E4DF}
\definecolor{wrongink}{HTML}{80362C}
\definecolor{rightbg}{HTML}{E0EEE3}
\definecolor{rightink}{HTML}{19523D}

\newtcolorbox{CasePanel}[1]{colback=#1,colframe=#1border,boxrule=.45bp,
  arc=2bp,left=4bp,right=4bp,top=3bp,bottom=3bp,boxsep=0bp,
  before skip=0bp,after skip=2bp}

\newcommand{\CaseNumbersOff}{\ifdefined\nolinenumbers\nolinenumbers\fi
}
\newcommand{\CaseNumbersLeft}[1]{\ifdefined\internallinenumbers
    \internallinenumbers
    \def\makeLineNumber{\hss\linenumberfont\LineNumber\hskip#1\relax}\fi
}
\newcommand{\CaseNumbersRight}{\ifdefined\internallinenumbers
    \internallinenumbers
    \def\makeLineNumber{\hskip254bp\relax\linenumberfont\LineNumber\hss}\fi
}
\newenvironment{CaseFigure}[3]{\par\noindent\begin{minipage}{\linewidth}\def\CaseCaption{#3}\CaseNumbersOff
  \fontsize{9}{10}\selectfont
  \begin{CasePanel}{case#1}\color{caseink}\CaseLabel{{\color{caseheading}\bfseries #2}}}{\end{CasePanel}\captionof{figure}{\CaseCaption}\end{minipage}\par
}

\newcommand{\CaseLabel}[1]{{\CaseNumbersLeft{25.7bp}\noindent #1\par}}
\newcommand{\CaseQuestion}[1]{\CaseLabel{\textbf{Question}\quad #1}}
\newcommand{\CaseChoices}[1]{\CaseLabel{#1}}
\newcommand{\Choice}[2]{\textbf{#1.}~#2\enskip}
\newcommand{\CaseViews}[1]{{\CaseNumbersOff\centering#1\par}\vspace{2bp}}

\newcommand{\CaseEvidenceColumn}[6]{\begin{minipage}[t]{.474\linewidth}
    \setlength{\parindent}{1pc}#6\noindent{\color{caseheading}\bfseries #1}\par
    \noindent\colorbox{#2}{\makebox[\dimexpr\linewidth-2\fboxsep\relax][l]{\small\color{#3}#4}}\par
    \vspace{2bp}#5\par
  \end{minipage}}

\newcommand{\CaseEvidence}[6]{{\CaseNumbersOff\noindent\colorbox{white}{\begin{minipage}{\dimexpr\linewidth-2\fboxsep\relax}
      \CaseEvidenceColumn{Overview evidence}{wrongbg}{wrongink}{#1\quad\textbf{#2}}{#3}{\CaseNumbersLeft{28.7bp}}\hfill\raisebox{-.75\baselineskip}{\small\(\longrightarrow\)}\hfill
      \CaseEvidenceColumn{JOINT/Crop evidence}{rightbg}{rightink}{#4\quad\textbf{#5}}{#6}{\CaseNumbersRight}\par
    \end{minipage}}\par}\vspace{2bp}}

\newcommand{\CaseCheck}[3]{{\small\setlength{\parindent}{1pc}\CaseNumbersLeft{25.7bp}\noindent\textbf{JOINT center check}\quad #1~#2\par
    #3\par
  }}
\newcommand{\CaseSpacer}{\vspace{5bp}}

%% file: 0_abstract.tex
\begin{abstract}
Vision-Language Models (VLMs) excel at visual understanding, but still struggle to reason reliably about spatial relationships. Models may capture relevant visual cues while failing to organize them into a coherent spatial representation, limiting their ability to reason reliably about spatial structure. To address this, we propose \textbf{Constella}, a post-training framework that strengthens spatial reasoning through configurational binding. Constella first constructs a privileged teacher context with a joint spatial view and correspondence-aware attention, which organizes local object evidence within a shared spatial frame. It then transfers this capability to the student through on-policy self-distillation, enabling the model to internalize the enhanced spatial reasoning while requiring only the original image at inference. To preserve general capabilities during spatial adaptation, we further introduce gradient decomposition to reduce interference between spatial learning and general capability retention. Finally, we present FineBind, a benchmark tailored for evaluating object-location association and global spatial configurations. Extensive experiments show that Constella consistently outperforms strong baselines across multiple spatial benchmarks with robust out-of-distribution generalization across diverse scenes.
\end{abstract}

%% file: 1_intro.tex
\begingroup
\setlength{\parindent}{1pc}
\setlength{\parskip}{0.6pt}
\setlength{\leftskip}{0pt}

\section{Introduction}
\noindent Vision-Language Models (VLMs) have gained strong visual perception capabilities~\cite{bai2025qwen3vl,zhu2025internvl3,huang2026step3vl}, enabling them to ground language in complex scenes. Yet, their spatial reasoning remains unreliable~\cite{jia2025omnispatial}. While recent efforts incorporate region-level or geometric cues to boost spatial awareness~\cite{chen2024spatialvlm,cheng2024spatialrgpt}, these methods often rely on specialized supervision or extra visual inputs that provide localized details at the expense of scene-level integration~\cite{qi2026beyondsemantics}. Effective spatial reasoning, however, demands balancing fine-grained local evidence with a globally consistent structural representation.

This limitation stems not from a fundamental lack of spatial knowledge, but from its fragmentation during visual encoding~\cite{yuan2026visionopd}. VLM encoders compress images into a limited set of tokens, often letting salient regions or background noise obscure subtle local details and sever their structural links to the broader scene~\cite{chen2025focusareas,hannan2026seeing}. Because reorganizing these fragmented cues in latent space is difficult to observe or supervise directly, we propose intervening prior to visual encoding. In particular, effective spatial reasoning requires a representation that preserves both local details and global structural alignment in a single view, a property we call configurational sufficiency.

To achieve configurational sufficiency without adding visual inputs at inference, we propose Constella, a framework for post training that distills privileged spatial reasoning into a single image model. Constella constructs a privileged teacher context by combining a JOINT spatial view with a crop of each object, using Correspondence Prior Attention (CPA) to align spatial evidence across views. Through self distillation on student rollouts, the student internalizes this structural reasoning using only the original image. To preserve general visual capabilities during spatial adaptation, we introduce gradient decomposition to suppress reduce conflicts between spatial learning and general visual skills. We also present FineBind, a spatial reasoning benchmark designed to evaluate object location association. Training on FineBind consistently improves performance across diverse external tasks, demonstrating broad generalization beyond the training distribution.

Our main contributions are summarized as follows:
\begin{list}{\arabic{enumi}.}{\usecounter{enumi}\setlength{\leftmargin}{1.25em}\setlength{\labelwidth}{0.75em}\setlength{\labelsep}{0.5em}\setlength{\topsep}{2.6pt}\setlength{\partopsep}{0pt}\setlength{\itemsep}{0pt}\setlength{\parsep}{0pt}}
\item We identify visual encoding as the primary bottleneck in multi object spatial reasoning, establishing the necessity of preserving global spatial layout and the fine visual details of the referred objects before latent compression.
\item We propose Constella, a post training framework that leverages privileged teacher context and visual correspondence to distill enhanced spatial reasoning into a single image model, combined with gradient decomposition to prevent performance degradation on general tasks throughout the process of spatial adaptation.
\item We present FineBind, a benchmark dedicated to spatial position evaluation. Constella achieves superior performance across multiple spatial tasks with robust out of distribution generalization across diverse scenes.
\end{list}
\endgroup

%% file: 2_related_work.tex
\section{Related Work}

\subsection{Vision-Language Models and Spatial Evidence}

Image--text training gave VLMs broad visual perception \cite{radford2021learning,jia2021scaling}. Flamingo~\cite{alayrac2022flamingo}, BLIP-2~\cite{li2023blip2}, and LLaVA~\cite{liu2023llava} connected visual encoders to language models to generate answers from images. Qwen2-VL~\cite{wang2024qwen2vl}, Qwen2.5-VL~\cite{bai2025qwen25vl}, and Qwen3-VL~\cite{bai2025qwen3vl} handle varied visual inputs. InternVL3~\cite{zhu2025internvl3} and InternVL3.5~\cite{wang2025internvl35} also strengthen visual perception and reasoning. These gains do not fully resolve spatial reasoning errors. An attention study~\cite{chen2025focusareas} finds that errors can coincide with attention to irrelevant objects. Beyond Semantics~\cite{qi2026beyondsemantics} shows that semantic signals can obscure positional cues, while Seeing Beyond Redundancy~\cite{hannan2026seeing} links visual token compression to task complexity. Penguin-VL~\cite{zhang2026penguinvl} preserves fine spatial cues, while Phi-4-reasoning-vision~\cite{aneja2026phi4} and Fine-R1~\cite{he2026finer1} strengthen perception and visual reasoning. Other methods add spatial supervision or geometric information. SpatialVLM~\cite{chen2024spatialvlm} uses 3D spatial data, and RoboSpatial~\cite{song2024robospatial} provides spatial training data for robotics. SpatialRGPT~\cite{cheng2024spatialrgpt} adds depth and region cues. SpatialGeo~\cite{guo2025spatialgeo} combines geometric and visual features, and GR3D~\cite{cheng2026gr3d} adds 3D grounding. Self-Evolving Spatial Reasoning~\cite{liu2026self} encourages consistent answers across geometric transformations. Language-conditioned grounding provides another way to connect object descriptions with image regions. MDETR~\cite{kamath2021mdetr} learns to localize regions from free-form text.  Constella focuses on how evidence for several objects is linked to their positions within one shared scene to support coherent spatial reasoning.

\subsection{On-Policy Distillation for Visual Reasoning}

Knowledge distillation transfers teacher predictions to a student \cite{hinton2015distilling}. LLAVADI~\cite{xu2024llavadi}, LLaVA-KD~\cite{cai2024llavakd}, and LLaVA-MoD~\cite{shu2024llavamod} extend this idea to multimodal models. VisPer-LM~\cite{jain2024visperlm} distills visual embeddings, while CompoDistill~\cite{kim2025compodistill} aligns visual attention. Training on fixed teacher outputs can leave a gap between training and student inference. On-Policy Distillation (OPD)~\cite{agarwal2024onpolicy} addresses this gap by supervising trajectories generated by the student. On-Policy Self-Distillation (OPSD)~\cite{zhao2026selfdistilled} uses different contexts to form a teacher and student within one model. Vision-OPD~\cite{yuan2026visionopd} gives the teacher a detailed crop. VA-OPD~\cite{liu2026visualadvantage} selects tokens that depend on visual evidence, while DOPD~\cite{yoon2026decomposed} steers learning toward visual grounding. VCSD~\cite{liang2026vcsd} and S$^2$VOPD~\cite{li2026s2vopd} use altered visual evidence to create training signals. VAD~\cite{zhang2026vad} isolates visual corrections, and OPD-V~\cite{aniri2026opdv} uses modality balance to select tokens. These methods strengthen visual supervision, but a relation among several objects also requires each local cue to be matched to its position in the same scene. Constella distills this shared spatial configuration into a student that operates solely on the original image. Reasoning post-training can also weaken general capabilities \cite{phan2025forgetting}; gradient decomposition aims to limit conflicting updates.

\subsection{Spatial Reasoning Benchmarks}

Spatial benchmarks differ in the evidence and viewpoint they require. What'sUp~\cite{kamath2023whatsup} holds object identity fixed while changing relations in controlled photographs. COCO-Spatial and GQA-Spatial from the same study use natural images, which contain more objects and familiar layout biases. Visual Spatial Reasoning (VSR)~\cite{liu2023vsr} broadens the range of relations tested in natural images through true-or-false judgments on spatial statements, including relations that depend on the reference frame. Spatial-MM~\cite{shiri2024spatialmm} tests object relations and questions with several reasoning steps, including questions from camera and human perspectives. Recent benchmarks probe further cases. CLEVR-3D-DeRef~\cite{martin2026clevr3dderef} asks models to resolve references using depth and relational context. SpatiaLQA~\cite{xie2026spatiallqa} tests spatial logic, while InfiniBench~\cite{wang2026infinibench} varies scene complexity. OmniSpatial~\cite{jia2025omnispatial} covers a broad range of spatial tasks. SpinBench~\cite{zhang2026spinbench} tests perspective and rotation, and CityCube~\cite{xu2026citycube} studies changes in viewpoint in urban scenes. OST-Bench~\cite{lin2025ostbench} evaluates spatial understanding during online exploration. MMSI-Bench~\cite{yang2026mmsibench} asks models to combine evidence across several images. These settings vary the evidence across views and time. FineBind focuses on several objects within one complex image. It is designed to evaluate
global configuration and object-to-position binding for multiple targets in complex scenes. For example, a single question can require comparing several candidate objects to identify a spatial extreme.

%% file: 3_methods.tex
\section{Methods}

\begin{figure*}[t]
    \centering
    \includegraphics[width=\textwidth]{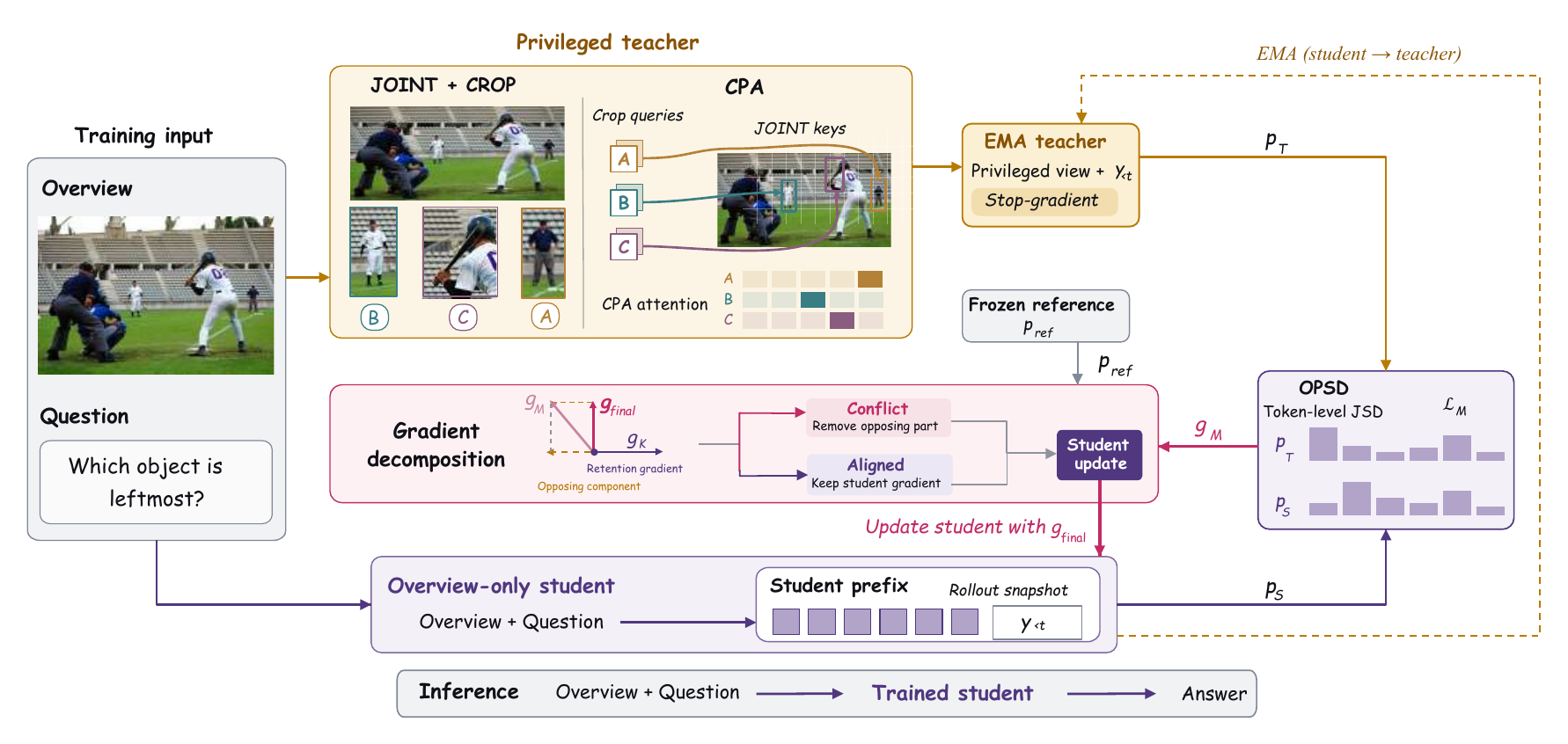}
    \caption{\textbf{Overview of the Constella architecture.} JOINT constructs a shared spatial view, and CPA grounds crop evidence into corresponding coordinates. An EMA teacher supervises the student model on sampled trajectories, while a frozen reference guides gradient decomposition to preserve general capabilities. Inference relies solely on the trained student.}
    \label{fig:constella-main}
\end{figure*}

\label{sec:method}

\subsection{Problem Setting and Framework}
\label{sec:method-setting}

We study spatial reasoning aimed at achieving configurational sufficiency. An instance from our target domain is represented as
\begin{equation}
(I^o,\mathcal C,q,y),\qquad
\mathcal C=(I_1^c,\ldots,I_K^c),
\end{equation}
where $I^o$ is an overview image, $\mathcal C$ contains $K$ object crops, $q$ is a multi-object question, and $y$ is its annotated answer. The student observes only $x_S=(I^o,q)$, whereas the Dense privileged teacher receives a direct sequence-level concatenation of the overview and all object crops, $x_T^D=[I^o;I_1^c;\cdots;I_K^c;q]$. Semicolons denote multimodal sequence order. Bounding boxes are used only to prepare the crops: neither box coordinates, $y$, nor the target relation are supplied to the model, and our objective contains no supervised cross-entropy on $y$. At inference, only $(I^o,q)$ is available to the student model for predicting relations.

Our framework, Constella, equips the privileged teacher with JOINT and correspondence-prior attention (CPA). JOINT restores a shared coordinate frame for the relevant objects; CPA binds crop-level identity evidence to the corresponding locations in that frame. The resulting teacher supervises an overview-only student on student-generated prefixes. A frozen overview-only reference additionally detects updates that conflict with the pretrained policy and corrects the distillation gradient. We maintain four model states: the trainable student $\theta$, a rollout snapshot $\theta^-$, an exponential-moving-average (EMA) teacher $\bar\theta$, and a frozen reference $\theta_{\mathrm{ref}}$.
Figure~\ref{fig:constella-main} summarizes the training and inference paths of the proposed framework for spatial reasoning.

\subsection{Teacher-Side JOINT and CPA}
\label{sec:method-cb}

\paragraph{JOINT.}
A frozen pixel relocalizer matches each crop back to the overview,
$(\widehat B_i,T_i,\rho_i)=\mathcal M(I_i^c,I^o)$, where $T_i$ maps crop pixels to overview coordinates, $\widehat B_i$ is the matched region, and $\rho_i$ is its reliability. JOINT and CPA are used only if every crop is localized reliably:
\begin{equation}
z=\prod_{i=1}^{K}\mathbf 1\!\left[
\rho_i\geq\eta_{\mathrm{loc}}
\ \land\
\operatorname{Valid}(T_i,\widehat B_i)\right].
\label{eq:method-gate}
\end{equation}
The gate depends only on pixel correspondence, not on the answer or predicted relation. For $z=1$, JOINT extracts a common scene window directly from the overview:
\begin{equation}
\begin{aligned}
B^J&=\operatorname{Clip}\!\left[
\operatorname{Expand}_{m}\!\left(
\operatorname{BBox}\!\left(\bigcup_i\widehat B_i\right)\right)\right],\\
I^J&=\mathcal R_J\!\left(\operatorname{Crop}(I^o,B^J)\right).
\end{aligned}
\label{eq:method-joint}
\end{equation}
Here $m$ is a fixed context margin and $\mathcal R_J$ is the actual resize-and-pad transform. If $A_J$ maps overview coordinates through this transform, the crop-to-JOINT correspondence is $F_i^J=A_J\circ T_i$. JOINT never pastes crop pixels: all targets remain in one scene-derived image and undergo the same transformation, preserving their relative layout while increasing their effective resolution.

When $z=0$, JOINT is not constructed and the teacher falls back to the Dense input. Accordingly, the routed teacher input is
\begin{equation}
x_T(z)=
\begin{cases}
[I^J;I_1^c;\cdots;I_K^c;q], & z=1,\\
x_T^D, & z=0,
\end{cases}
\label{eq:method-routing}
\end{equation}
and CPA is disabled on this Dense fallback route.

\paragraph{Correspondence-prior attention.}
Let $\mathcal U_i$ and $\mathcal P_J$ be the content-token indices of crop $i$ and JOINT, respectively. A frozen feature extractor gives a pooled crop descriptor $\bar c_i$ and JOINT token features $o^J(p)$. We combine their cosine similarity $s_i(p)=\cos(\bar c_i,o^J(p))$ with the recovered geometry. Specifically, let $r_p$ be the normalized spatial center of JOINT token $p$, and let $\Omega_i^J$ be the normalized image of $\widehat B_i$ under $A_J$. CPA defines
\begin{equation}
G_i(p)=
\frac{\exp\!\left[\frac{s_i(p)}{\tau}-\frac{d(r_p,\Omega_i^J)^2}{2\sigma_i^2}\right]}
{\sum_{p'\in\mathcal P_J}\exp\!\left[\frac{s_i(p')}{\tau}-\frac{d(r_{p'},\Omega_i^J)^2}{2\sigma_i^2}\right]},
\label{eq:method-prior}
\end{equation}
where $d$ is the distance to the continuous matched region, $\tau$ is an appearance temperature, and $\sigma_i$ is a spatial bandwidth.

At fixed layer--head pairs $\mathcal A_{\mathrm{CPA}}$, CPA changes only crop-query-to-JOINT-key logits:
\begin{equation}
\begin{aligned}
\widetilde A_{ab}
&=\operatorname{softmax}_{b}\!\left(
\frac{Q_aK_b^\top}{\sqrt{d_h}}+M_{ab}+B_{ab}\right),\\
B_{ab}
&=\mathbf 1[a\!\in\!\mathcal U_i,b\!\in\!\mathcal P_J]
\lambda_A\log(G_i(b)+\epsilon_A).
\end{aligned}
\label{eq:method-cpa-attention}
\end{equation}
The bias is zero outside $\mathcal A_{\mathrm{CPA}}$ and whenever $z=0$; the original attention mask $M$ is unchanged. Because JOINT precedes the crops in the serialized input, crop queries can retrieve scene evidence from their matched JOINT locations. Thus, JOINT supplies a relation-bearing configuration, while CPA associates crop-specific identity with positions inside it. Both modules are training-free and introduce no optimized parameters.

\subsection{On-Policy Self-Distillation}
\label{sec:method-distillation}

The rollout snapshot samples an overview-only response
\begin{equation}
\hat y\sim\pi_{\theta^-}(\cdot\mid x_S).
\label{eq:method-rollout}
\end{equation}
At every sampled prefix $\hat y_{<t}$, student and teacher distributions are
\begin{equation}
\begin{aligned}
p_{S,t}&=\pi_\theta(\cdot\mid x_S,\hat y_{<t}),\\
p_{T,t}&=\operatorname{sg}\!\left[
\pi_{\bar\theta}^{B}(\cdot\mid x_T(z),\hat y_{<t})\right],
\end{aligned}
\label{eq:method-shared-prefix}
\end{equation}
where $B$ denotes the CPA biases and $\operatorname{sg}$ stops gradients. Hence the teacher corrects states actually visited by the student rather than fixed expert prefixes.

For efficiency, we retain the student's stopped-gradient top-$K_d$ support $\mathcal S_t$ and aggregate all remaining vocabulary mass into a shared tail symbol $\bot$:
$\widetilde p_t(v)=p_t(v)$ for $v\in\mathcal S_t$ and
$\widetilde p_t(\bot)=1-\sum_{v\in\mathcal S_t}p_t(v)$.
With $m_t=\beta\widetilde p_{T,t}+(1-\beta)\widetilde p_{S,t}$, the token loss is
\begin{equation}
\begin{aligned}
\mathcal J_t &= \operatorname{JSD}_\beta(\widetilde p_{T,t}, \widetilde p_{S,t}) \\
&= \beta D_{\mathrm{KL}}(\widetilde p_{T,t}\|m_t) + (1-\beta)D_{\mathrm{KL}}(\widetilde p_{S,t}\|m_t).
\end{aligned}
\label{eq:method-jsd}
\end{equation}
For a logical batch $\mathcal B$, a fixed mask $w_{n,t}$ selects valid response tokens, $Z_n=\sum_t w_{n,t}$, and $\mathcal B_+=\{n:Z_n>0\}$. We optimize the trajectory-balanced objective
\begin{equation}
\mathcal L_M=
\frac{1}{|\mathcal B_+|}\sum_{n\in\mathcal B_+}
\frac{1}{Z_n}\sum_t w_{n,t}\mathcal J_{n,t}.
\label{eq:method-main-loss}
\end{equation}
Sampling, support selection, and teacher predictions are held fixed within the update.

\subsection{KL-Guided Gradient Correction}
\label{sec:method-gradient}

We use a frozen reference to identify updates that conflict with retention of the pretrained policy. The reference receives the same context as the student,
$p_{\mathrm{ref},n,t}=\operatorname{sg}[\pi_{\theta_{\mathrm{ref}}}(\cdot\mid x_{S,n},\hat y_{n,<t})]$.
The full-vocabulary retention objective is
\begin{equation}
\mathcal L_K^{\mathrm{full}}=
\frac{\lambda_{\mathrm{KL}}}{Z_K}
\sum_{n,t}a_{n,t}
D_{\mathrm{KL}}(p_{S,n,t}\|p_{\mathrm{ref},n,t}),
\label{eq:method-full-kl}
\end{equation}
where $a_{n,t}$ selects retention positions and $Z_K$ is a fixed batch normalizer. For the sampled token $v_{n,t}=\hat y_{n,t}$, let $q_{n,t}$ be the rollout sampling distribution, $R_{n,t}=p_{S,n,t}(v_{n,t})/q_{n,t}(v_{n,t})$, and $f_{n,t}=\log[p_{S,n,t}(v_{n,t})/p_{\mathrm{ref},n,t}(v_{n,t})]$. The sampled-token surrogate is
\begin{equation}
\mathcal L_K =
\frac{\lambda_{\mathrm{KL}}}{Z_K}\sum_{n,t}a_{n,t}
\operatorname{sg}\!\left[R_{n,t}(f_{n,t}-b_{n,t})\right]
\log p_{S,n,t}(v_{n,t}),
\label{eq:method-kl-surrogate}
\end{equation}
where $b_{n,t}$ is a baseline that may depend on the prefix. At a fixed prefix, the gradient is unbiased for the KL gradient with full sampling support, exact importance weights, and position selection and baselines independent of the current token draw.

We use the retention gradient to correct conflicting distillation updates. Let
$g_M=\nabla_\theta\mathcal L_M$, $g_K=\nabla_\theta\mathcal L_K$,
$\gamma=\langle g_M,g_K\rangle$, and $\kappa=\|g_K\|_2^2$. We apply the asymmetric correction
\begin{equation}
g_{\mathrm{final}}=
g_M-\mathbf 1[\gamma<0]\frac{\gamma}{\kappa+\varepsilon_p}g_K.
\label{eq:method-correction}
\end{equation}
For $\gamma<0$, the correction reduces the conflict to $\langle g_{\mathrm{final}},g_K\rangle=\gamma\varepsilon_p/(\kappa+\varepsilon_p)$. We compute the correction on globally accumulated, unscaled gradients before clipping and the AdamW step, reducing dot products across parameter shards.

After the student update, the teacher is refreshed by
$\bar\theta\leftarrow\mu\bar\theta+(1-\mu)\theta$.
At inference, we discard JOINT, CPA, the EMA teacher, and the reference policy. The deployed model is therefore the original overview-only student and requires no crops, boxes, relocalization, or attention intervention when answering spatial questions.

\begin{figure*}[t]
  \centering
  \includegraphics[width=\textwidth]{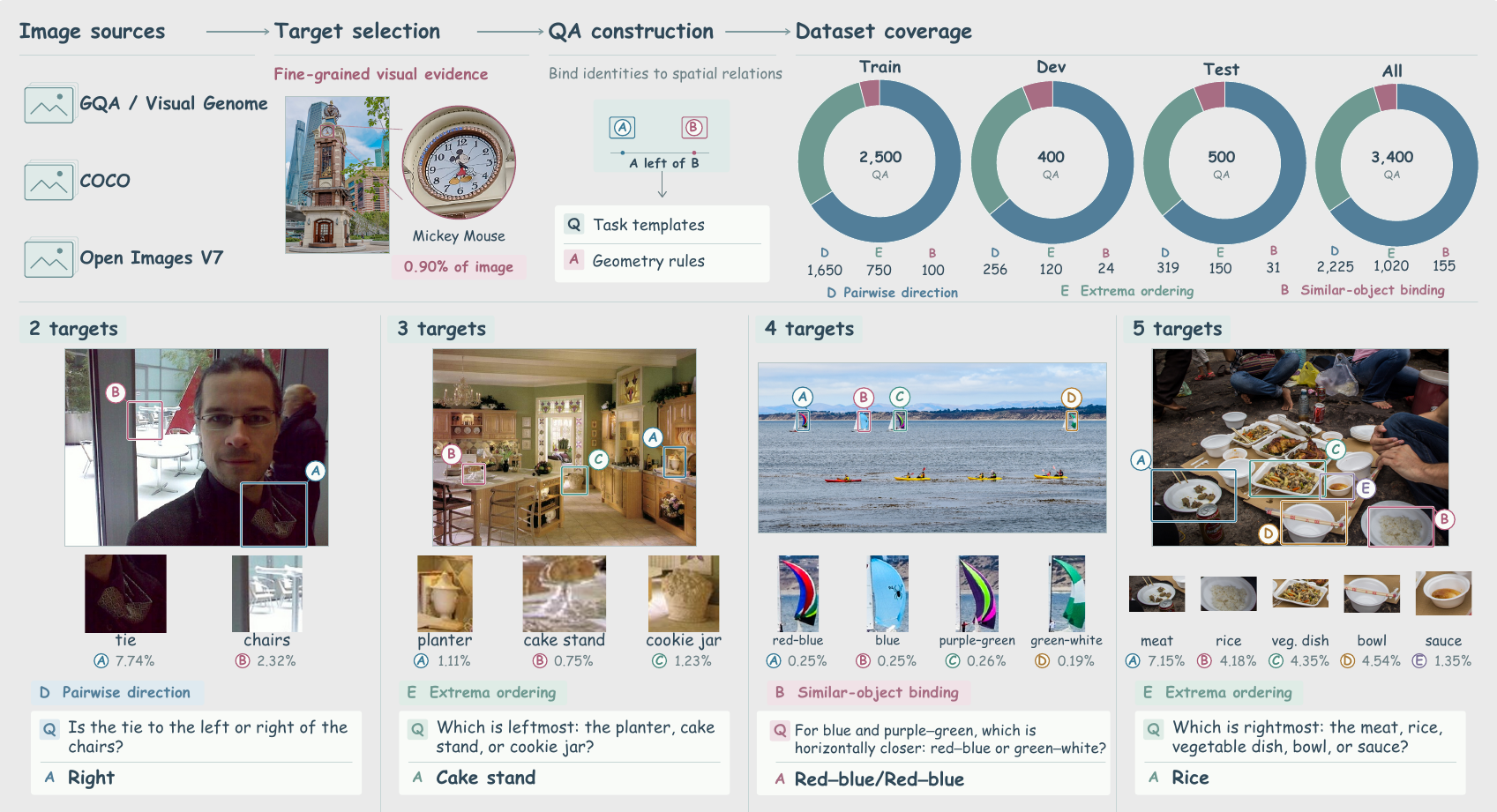}
  \caption{\textbf{FineBind construction, task distribution, and qualitative examples..} (Top) We select target objects in natural images, generate spatial questions from task templates, and compute ground-truth answers from bounding-box geometry. The charts show the distribution of 3,400 question–answer pairs by task type and split. (Bottom) Scenes with two to five targets illustrate pairwise direction, extrema ordering, and similar-object binding. The percentages indicate each target box’s share of the image area. At evaluation, the model receives the original image and question.}
  \label{fig:finebind-teaser}
\end{figure*}

\subsection{Data Construction}
\label{sec:method-finebind}

To systematically evaluate configurational spatial reasoning, we introduce \textbf{FineBind}, a dedicated benchmark constructed from natural scenes in GQA/Visual Genome~\cite{hudson2019gqa,krishna2017visualgenome}, COCO~\cite{lin2014coco}, and Open Images V7~\cite{kuznetsova2020openimages}. Each instance targets scenes containing two to five co-occurring objects, with target counts uniformly distributed across the dataset to balance spatial complexity. We instantiate programmatic query templates covering three core spatial dimensions: pairwise directional orientation, spatial extrema ordering, and object-location association. Ground-truth target relations are computed deterministically from gold bounding geometry, while target bounding boxes supply object crops for the privileged teacher context during post-training.

To guarantee high label fidelity and eliminate spatial ambiguity, we implement a dual-stage quality verification pipeline combining automated heuristic filtering with human expert auditing. Automatically, candidates undergo algorithmic checks for object visibility, geometric validity, referring expression uniqueness, and crop correspondence. Subsequently, human annotators independently verify the candidate samples to filter out visually ambiguous referring expressions, occluded targets, or boundary spatial cases. 
We partition the curated dataset by grouping instances strictly by source image family to prevent data contamination across splits, keeping horizontal-mirror variants within their parent image split. The resulting dataset comprises 2,500 training, 400 validation, and 500 test samples.

%% file: 4_experiments.tex
\section{Experiments}

\begingroup
\renewcommand{\arraystretch}{1.08}

\subsection{Experimental Setup}
\label{main_setup}

\noindent\textbf{Datasets.}
We train on 2,500 FineBind questions and evaluate on its test set of 500 questions, fixed before the student pilots. FineBind is a fine-grained spatial reasoning dataset with two to five referred objects per question, uniformly represented across these counts. Its tasks cover pairwise direction, extrema ordering, and disambiguation of visually similar objects. We assess transfer on nine external benchmarks: COCO-Spatial and GQA-Spatial~\cite{kamath2023whatsup}, MultiHopSpatial~\cite{multihop}, OST-Bench~\cite{lin2025ostbench}, SpatialMQA~\cite{spatialmqa}, Spatial-MM~\cite{shiri2024spatialmm}, SpatiaLQA~\cite{xie2026spatiallqa}, SPINBench~\cite{zhang2026spinbench}, and COMFORT~\cite{comfort}. These benchmarks span object relations, compositional reasoning, reference frames, and scene understanding across a range of visual contexts.

\noindent\textbf{Baselines.}
Tables~\ref{tab:main-a} and~\ref{tab:main-b} compare Constella with its starting models, Qwen3.5-4B-Base and Qwen3.5-9B-Base, as well as general VLMs and spatial reasoning methods. Comparisons with the starting models quantify the gains from post-training on the same model backbones.

\noindent\textbf{Implementation.}
We train the 9B model on eight NVIDIA A100-SXM4-40GB GPUs with BF16 and a global batch size of 96. The learning rate is $2\times10^{-6}$, with 10 warmup steps followed by a constant schedule. Weight decay is 0.01. The teacher uses JOINT with CPA when the correspondence gate passes and Dense crop concatenation otherwise. CPA applies a bias of 0.05 at language layers 3 and 7. Distillation uses the top 100 tokens and a shared tail, an EMA teacher, and gradient correction guided by KL to stabilize the student updates.

\noindent\textbf{Evaluation.}
All student evaluations use the original visual input and question, without JOINT, object crops, or CPA. We retain each benchmark\textquotesingle s scoring convention and report scores as percentages and improvements in percentage points. Evaluation of general capabilities uses the decoding settings in Table~\ref{tab:general}. Table~\ref{tab:ablation} reports direct evaluation with privileged views for comparison with standard student inference.

\begin{table*}[t]
    \centering
    \caption{Main results on FineBind and four external spatial reasoning benchmarks. Scores are percentages.}
    \label{tab:main-a}
    \small
    \setlength{\tabcolsep}{2.7pt}
    \begin{tabularx}{\linewidth}{@{}Xrrrrr@{}}
        \toprule
        Method
        & FineBind
        & OST-Bench
        & COMFORT
        & \shortstack{MultiHop\\Spatial}
        & Spatial-MM \\
        \midrule

        \rowcolor{bindQwen}
        \multicolumn{6}{c}{\itshape Qwen3.5 Backbones} \\
        Qwen3.5-4B-Base
        & 45.80 & 42.25 & 66.37 & 40.98 & 63.26 \\
        Qwen3.5-9B-Base
        & 53.20 & 43.19 & 72.29 & 38.73 & 63.15 \\
        \midrule

        \rowcolor{bindOther}
        \multicolumn{6}{c}{\itshape Other Open-source Backbones} \\
        InternVL-2.5-8B
        & 37.20 & 39.11 & 74.33 & 36.82 & 65.18 \\
        LLaVA-OV-7B
        & 50.80 & 38.03 & 68.37 & 36.56 & 60.76 \\
        \midrule

        \rowcolor{bindSpatial}
        \multicolumn{6}{c}{\itshape Spatial Reasoning Methods} \\
        SpatialLadder-3B
        & 45.40 & 36.88 & 73.78 & 29.89 & 59.58 \\
        GeoAnchor-2B
        & 53.40 & 28.15 & 46.11 & 24.82 & 42.97 \\
        GeoThinker-7B
        & 51.00 & 25.96 & 69.91 & 30.69 & 53.30 \\
        HATCH-3B
        & 49.00 & 30.14 & 73.08 & 27.00 & 36.48 \\
        MindCube-RL-3B
        & 43.80 & 35.82 & 74.22 & 33.89 & 35.68 \\
        ViewFusion-4B
        & 50.00 & 42.12 & 62.69 & 38.42 & 62.03 \\
        VLM3R-7B
        & 49.60 & 39.57 & 65.05 & 38.80 & 55.86 \\
        \midrule

        \rowcolor{bindOurs}
        \multicolumn{6}{c}{\itshape Ours} \\
        Constella-4B
        & 62.40 & 43.72 & 74.50 & 41.96 & 69.06 \\
        Constella-9B
        & \textbf{66.20}
        & \textbf{44.36}
        & \textbf{74.81}
        & \textbf{43.87}
        & \textbf{70.93} \\
        \bottomrule
    \end{tabularx}

    \vspace{3pt}
    \parbox{\linewidth}{\fontsize{8}{9.3}\selectfont
        \textit{Note.}
        Model sizes denote the nominal evaluated versions and may exclude
        additional visual or geometric modules.
        GeoThinker uses the 7B Scaled+VGGT variant.
        VLM3R uses a 7B backbone with LoRA and CUT3R.
    }
\end{table*}

\subsection{Main Results}

\noindent\textbf{Spatial performance and transfer.}
Tables~\ref{tab:main-a} and~\ref{tab:main-b} show that Constella improves over its starting models on FineBind and all nine external benchmarks at both scales, using only 2,500 training questions. Constella-9B achieves the highest score among the evaluated models on every benchmark, exceeding the strongest non-Constella result by 12.80 points on FineBind. Transfer is broad but nonuniform: relative to the 9B starting model, the largest external gain is 19.02 points on SpatiaLQA, followed by 7.78 on Spatial-MM. Improvements also extend to COMFORT, SPINBench, and OST-Bench, which involve different reference frames or observation contexts. COCO-Spatial and GQA-Spatial show smaller absolute gains from starting accuracies above 93\%.

\noindent\textbf{Benefits at both model scales.}
Constella-4B outperforms Qwen3.5-9B-Base on nine of the ten evaluations and matches it on GQA-Spatial. Thus, the benefits of spatial post-training extend beyond those obtained by increasing the starting model from 4B to 9B. Constella-9B nevertheless outperforms Constella-4B on every benchmark, showing that the larger model also benefits from the same training approach.

\begin{table*}[t]
    \centering
    \caption{Main results on five additional spatial reasoning benchmarks. Scores are percentages.}
    \label{tab:main-b}
    \small
    \setlength{\tabcolsep}{2.7pt}
    \begin{tabularx}{\linewidth}{@{}Xrrrrr@{}}
        \toprule
        Method
        & SpatialMQA
        & SpatiaLQA
        & COCO-Spatial
        & GQA-Spatial
        & SPINBench \\
        \midrule

        \rowcolor{bindQwen}
        \multicolumn{6}{c}{\itshape Qwen3.5 Backbones} \\
        Qwen3.5-4B-Base
        & 42.75 & 44.70 & 91.51 & 94.69 & 62.21 \\
        Qwen3.5-9B-Base
        & 41.82 & 48.05 & 93.34 & 95.31 & 59.91 \\
        \midrule

        \rowcolor{bindOther}
        \multicolumn{6}{c}{\itshape Other Open-source Backbones} \\
        InternVL-2.5-8B
        & 31.78 & 35.34 & 93.15 & 94.49 & 57.61 \\
        LLaVA-OV-7B
        & 32.62 & 43.65 & 92.33 & 92.56 & 56.12 \\
        \midrule

        \rowcolor{bindSpatial}
        \multicolumn{6}{c}{\itshape Spatial Reasoning Methods} \\
        SpatialLadder-3B
        & 29.74 & 45.67 & 84.96 & 85.04 & 50.42 \\
        GeoAnchor-2B
        & 30.67 & 56.56 & 74.69 & 73.40 & 48.08 \\
        GeoThinker-7B
        & 37.36 & 20.60 & 93.67 & 94.42 & 54.18 \\
        HATCH-3B
        & 28.35 & 51.77 & 92.22 & 91.45 & 48.52 \\
        MindCube-RL-3B
        & 20.07 & 46.91 & 88.72 & 89.94 & 49.18 \\
        ViewFusion-4B
        & 35.78 & 62.25 & 92.97 & 94.69 & 58.38 \\
        VLM3R-7B
        & 34.48 & 47.71 & 91.89 & 94.21 & 55.09 \\
        \midrule

        \rowcolor{bindOurs}
        \multicolumn{6}{c}{\itshape Ours} \\
        Constella-4B
        & 47.39 & 60.66 & 94.19 & 95.31 & 62.40 \\
        Constella-9B
        & \textbf{48.05}
        & \textbf{67.07}
        & \textbf{94.64}
        & \textbf{96.07}
        & \textbf{63.60} \\
        \bottomrule
    \end{tabularx}
\end{table*}

\subsection{Privileged Evidence and Student Transfer}

Table~\ref{tab:ablation} separates direct evaluation of the starting weights under different visual configurations from distillation into students evaluated with the overview only. The two stages test whether the privileged configuration improves predictions and whether its benefit transfers to the student.

\begin{table*}[t]
    \centering
    \caption{Direct evaluation and student transfer at the 4B and 9B scales. Scores are accuracy (\%). The upper block evaluates the starting weights under different visual configurations. In the lower block, parentheses identify the teacher configuration. All students use the overview only at evaluation.}
    \label{tab:ablation}
    \small
    \setlength{\tabcolsep}{2.7pt}
    \begin{tabular*}{\linewidth}{@{\extracolsep{\fill}}lrrrr@{}}
        \toprule
        Method
        & FineBind
        & SpatialMQA
        & Spatial Lab
        & Spatial-MM \\
        \midrule

        \multicolumn{5}{@{}l}{\itshape Direct evaluation of the starting model
        } \\
        Qwen3.5-4B-Base
        & 45.80 & 42.75 & 42.50 & 63.26 \\
        + JOINT
        & 50.00 & 42.94 & 45.00 & 63.37 \\
        + JOINT + CPA
        & 90.00 & 44.70 & 45.43 & 65.76 \\
        Qwen3.5-9B-Base
        & 53.20 & 41.82 & 45.43 & 63.15 \\
        + JOINT
        & 53.60 & 43.59 & 44.29 & 63.26 \\
        + JOINT + CPA
        & 63.40 & 48.05 & 44.64 & 64.32 \\
        \midrule

        \multicolumn{5}{@{}l}{\itshape Students using the overview after distillation
        } \\
        Constella-4B (JOINT)
        & 60.00 & 42.19 & 45.43 & 65.76 \\
        Constella-4B (JOINT + CPA)
        & 62.40 & 47.39 & 47.93 & 69.06 \\
        Constella-9B (JOINT)
        & 55.80 & 44.80 & 49.64 & 68.37 \\
        Constella-9B (JOINT + CPA)
        & 66.20 & 48.05 & 49.93 & 70.93 \\
        \bottomrule
    \end{tabular*}
\end{table*}

\noindent\textbf{Direct benefit of privileged views.}
JOINT alone improves seven of the eight comparisons across scales and benchmarks, although the gains are generally small and Spatial Lab decreases by 1.14 points at 9B. JOINT+CPA improves all eight comparisons over JOINT, with the largest gains on FineBind. The full configuration therefore provides a stronger source of teacher predictions than JOINT alone, while its advantage varies substantially across benchmarks and model scales for subsequent student distillation.

\noindent\textbf{Transfer to students using the overview.}
Using JOINT+CPA instead of JOINT for distillation improves all eight student comparisons, including a gain of 5.20 points on SpatialMQA at 4B. All students receive the same overview image and question at evaluation. The comparison varies the teacher configuration used during distillation.

\noindent\textbf{Teacher accuracy and student gains.}
FineBind reveals a reversal across scales. Switching from JOINT to JOINT+CPA improves the initial teacher by 40.00 points at 4B and 9.80 at 9B, whereas the student gains are 2.40 and 10.40 points, respectively. A larger direct teacher gain therefore does not necessarily yield a larger student improvement. Direct evaluation measures predictions from the starting weights. Distillation uses token distributions along student rollouts and updates the teacher through EMA.

\subsection{Efficiency and General Capabilities}

\noindent\textbf{Cost of privileged views.}
Table~\ref{tab:runtime} estimates the time for 1,000 serial questions on one A100. JOINT increases runtime over Overview by 18.8\% at 4B and 24.8\% at 9B, while adding CPA increases runtime over JOINT by only 0.4\% and 2.4\%, respectively. Most of the measured overhead is therefore associated with the JOINT configuration. These costs characterize direct inference with privileged views. Deployed students do not invoke these operations.

\begin{table}[t]
    \centering
    \caption{Projected time for 1,000 serial questions on one A100 GPU. Measurements use the starting weights, include image reading, preprocessing, and generation, and exclude model loading. They do not measure the latency of trained students.}
    \label{tab:runtime}
    \small
    \setlength{\tabcolsep}{2.7pt}
    \begin{tabular*}{\linewidth}{@{\extracolsep{\fill}}llrr@{}}
        \toprule
        \shortstack{Starting\\model}
        & Input
        & \shortstack{Time for\\1,000 (s)}
        & \shortstack{vs.\\Overview} \\
        \midrule

        Qwen3.5-4B-Base
        & Overview & 195.16 & -- \\
        & JOINT & 231.88 & +18.8\% \\
        & JOINT + CPA & 232.82 & +19.3\% \\
        \midrule

        Qwen3.5-9B-Base
        & Overview & 205.76 & -- \\
        & JOINT & 256.86 & +24.8\% \\
        & JOINT + CPA & 262.90 & +27.8\% \\
        \bottomrule
    \end{tabular*}
\end{table}

\noindent\textbf{Changes in general capabilities.}
Table~\ref{tab:general} shows that spatial gains do not imply uniform retention of general capabilities. Constella-9B declines on all four evaluations. At 4B, GSM8K, MATH-500, and AIME2024 improve by 6.90, 17.20, and 31.97 percentage points, respectively. Bamboogle drops by 14.56 points at 4B and 12.01 at 9B despite consistent improvements on the spatial benchmarks. The shared decline on Bamboogle marks a limitation of the current training recipe, while the mathematical results depend on model scale under the same spatial training framework.

\begin{table}[t]
    \centering
    \caption{General mathematical and reasoning capabilities after spatial post-training. All scores are percentages. Higher is better and best results are bold. GSM8K, MATH-500, and Bamboogle use greedy pass@1. AIME2024 uses avg@4 over four samples with temperature 0.6.}
    \label{tab:general}
    \small
    \setlength{\tabcolsep}{2pt}
    \begin{tabular*}{\linewidth}{@{\extracolsep{\fill}}lrrrr@{}}
        \toprule
        Method
        & GSM8K
        & \shortstack{MATH-\\500}
        & \shortstack{AIME\\2024}
        & Bamboogle \\
        \midrule

        Qwen3.5-9B-Base
        & \textbf{96.51}
        & \textbf{91.40}
        & 41.67
        & \textbf{47.20} \\
        ViewFusion-4B
        & 91.96 & 86.20 & 30.00 & 15.20 \\
        Qwen3.5-4B-Base
        & 86.50 & 71.00 & 11.33 & 34.40 \\
        HATCH-3B
        & 82.41 & 63.20 & 4.17 & 4.80 \\
        InternVL-2.5-8B
        & 82.26 & 54.40 & 2.50 & 18.40 \\
        SpatialLadder-3B
        & 77.63 & 52.60 & 0.83 & 6.40 \\
        LLaVA-OV-7B
        & 75.51 & 45.80 & 0.83 & 11.20 \\
        VLM3R-7B
        & 69.90 & 44.80 & 0.83 & 10.40 \\
        MindCube-RL-3B
        & 73.01 & 44.60 & 0.00 & 6.40 \\
        GeoThinker-7B
        & 39.42 & 25.00 & 0.00 & 8.80 \\
        GeoAnchor-2B
        & 3.49 & 8.20 & 0.00 & 1.60 \\
        \midrule

        Constella-4B
        & 93.40 & 88.20 & \textbf{43.30} & 19.84 \\
        Constella-9B
        & 93.86 & 85.00 & 33.33 & 35.19 \\
        \bottomrule
    \end{tabular*}
\end{table}

\endgroup

%% file: 5_conclusion.tex
\section{Conclusion}

We introduced Constella to improve spatial reasoning by keeping object evidence and relative layout connected. JOINT builds a shared view of the targets, and CPA links individual crops to locations in that view. On-policy self-distillation transfers this privileged teacher signal to a student that receives only the original image and question at inference. We also introduced FineBind to test spatial binding in scenes with multiple objects. Constella-9B gains 13.00 points on FineBind over its starting model, improves on nine external spatial benchmarks. Ablations show that CPA adds value to JOINT and that this advantage can reach overview-only students. General capability changes remain mixed, and our evaluation focuses on single-image questions with object regions available during training. These limits motivate more direct tests of binding errors and capability retention.

%% file: supplement.tex
\captionsetup{hypcap=false}
\raggedbottom
\setlength{\multicolsep}{6pt plus 2pt minus 1pt}

\begin{multicols}{2}
\section*{Appendix}
\input{supplement/sections/dataset}
\input{supplement/sections/training}
\input{supplement/sections/derivations}
\input{supplement/sections/cost}

\end{multicols}
\input{supplement/sections/cases}

%% file: supplement/sections/dataset.tex
This supplement documents the evaluation data, analyzes geometric robustness and answer transitions, and gives the training settings and supporting derivations. Qualitative cases are in Sec.~\ref{sec:qualitative}.

\section{Dataset}
\subsection{Training dataset}
\textbf{Source images.} FineBind uses natural photographs from GQA/Visual Genome~\cite{hudson2019gqa}, COCO, and Open Images V7. Horizontal mirrors stay in the family of their parent image. Section~\ref{sec:data-generation} describes label and crop construction.

\smallskip
\textbf{Training inputs.} FineBind has 2,500 training, 400 development, and 500 test questions. The test set was fixed before the student pilots. Its design targets even coverage of questions referring to two through five objects. At evaluation the student receives the original image and question.

\smallskip
\textbf{Task types.} Pairwise direction compares two targets along an image axis. Extrema asks for the target farthest in the stated direction. Same- or near-category binding identifies visually similar instances from their attributes before comparing positions.

\smallskip
\textbf{Split audit.} Archived audit records report zero overlap for overview hashes, source image IDs, image families, and counterfactual groups. These keys cover reused source images and their known variants. Distinct photographs of the same physical scene require a separate scene-level audit.

\subsection{Scoring conventions}
The archived FineBind evaluator ranks candidate labels at a common prefix or scores the parsed greedy answer. The two readouts can differ when the generated response includes an explanation. Archived qualitative predictions use greedy decoding. External runs use each benchmark's scoring convention.

\smallskip
\textbf{General capability tests.} Main Table 5 reports greedy pass@1 for GSM8K, MATH-500, and Bamboogle. For AIME2024, the table reports avg@4 from four responses sampled at a temperature of 0.6.

\subsection{Spatial benchmarks}
Counts denote the archived evaluation sets. They can differ from the size of a benchmark's full release.

\smallskip
\textbf{OST-Bench}~\cite{lin2025ostbench} tests online spatio-temporal reasoning from observations available at the queried time.

\textbf{COMFORT}~\cite{comfort} tests spatial reference frames. The archived setting uses English hard yes/no questions for positive spatial relations (\(N=12{,}240\)).

\smallskip
\textbf{MultiHopSpatial}~\cite{multihop} tests spatial composition across relational steps (4,500 questions).

\smallskip
\textbf{Spatial-MM}~\cite{shiri2024spatialmm} covers spatial questions in multiple-choice and open-text formats. The archived run contains 1,556 multiple-choice and 322 open-text items.

\smallskip
\textbf{SpatialMQA}~\cite{spatialmqa} measures image-grounded spatial relations (1,076 examples).

\smallskip
\textbf{SpatiaLQA}~\cite{xie2026spatiallqa} tests spatial logical reasoning, including questions requiring distinct spatial facts to be combined.

\smallskip
\textbf{COCO-Spatial and GQA-Spatial}~\cite{kamath2023whatsup} test object relations in real images. The archived comparison contains 2,687 COCO-Spatial and 1,451 GQA-Spatial examples.

\smallskip
\textbf{SPINBench}~\cite{zhang2026spinbench} tests perspective and rotation (2,739 questions). The 1,400-example archived \textbf{Spatial Lab} run informs the teacher-view analysis in main Table 3.

\section{Additional Diagnostics}
\subsection{Sensitivity to localization error}
For all 30 curated three-object examples, we recomputed the winning target from the box centers. Let \(\delta\) be the winner-to-runner-up gap divided by image width. Under independent center shifts of at most \(e\) image widths, the answer stays fixed for \(\delta>2e\). Otherwise, the two closest targets can tie or switch order.

\begin{center}
\begin{minipage}{\columnwidth}
  \captionof{table}{\textbf{Ordering under bounded center shifts.} Counts are out of 15 per query direction.}
  \label{tab:center-shifts}
  \small
  \begin{tabular*}{\linewidth}{@{\extracolsep{\fill}}lrrr@{}}
    \toprule
    Query & \(e=5\%\) & \(e=10\%\) & \(e=15\%\) \\
    \midrule
    Leftmost & 15 & 12 & 8 \\
    Rightmost & 15 & 14 & 5 \\
    \bottomrule
  \end{tabular*}
\end{minipage}
\end{center}

All labels match box-center order. The median radius \(\delta/2\) is 14.35\% of image width. The smallest, 6.34\%, occurs for a beak and a branch. These visualization examples were screened for target spacing. The diagnostic tests position errors with identities fixed.

\smallskip
\textbf{What the radius measures.} A common center shift cancels in the difference. Independent shifts can close the gap by \(2eW\) for overview width \(W\). In the beak--branch example, opposite shifts of about 6.34\% of \(W\) can create a tie even with both targets correctly identified.

%% file: supplement/sections/training.tex
\subsection{Bounds on paired corrections}
On the same 500 FineBind test questions, let \(a,b\) be the numbers of correct answers before and after training. Let \(R\) count wrong-to-right answers and \(H\) count right-to-wrong answers. Then \(R-H=b-a\). The feasible range for \(R\) is \([\max(0,b-a),\min(b,500-a)]\).

\begin{center}
\begin{minipage}{\columnwidth}
  \captionof{table}{\textbf{Feasible transition counts} from the FineBind scores in main Table 1.}
  \label{tab:paired-corrections}
  \small
  \begin{tabular*}{\linewidth}{@{\extracolsep{\fill}}lrrr@{}}
    \toprule
    Scale & Net \(R-H\) & Corrections \(R\) & Regressions \(H\) \\
    \midrule
    4B & 83 & 83--271 & 0--188 \\
    9B & 65 & 65--234 & 0--169 \\
    \bottomrule
  \end{tabular*}
\end{minipage}
\end{center}

The gain fixes the minimum number of corrected items. Identifying which relations are repaired or newly missed requires matching per-item predictions.

\smallskip
The overlap \(I\) of the two sets of correct predictions satisfies \(\max(0,a+b-500)\le I\le\min(a,b)\). For any feasible \(I\), \(R=b-I\) and \(H=a-I\).

\section{Training Details}
\begin{center}
\begin{minipage}{\columnwidth}
  \captionof{table}{\textbf{9B implementation settings.}}
  \label{tab:training-settings}
  \small
  \begin{tabularx}{\linewidth}{@{}l>{\raggedright\arraybackslash}X@{}}
    \toprule
    Setting & Value \\
    \midrule
    Compute and precision & \(8\times\) A100-SXM4-40GB, BF16 \\
    Global / per-device batch & 96 / 1 per microbatch \\
    Learning rate & \(2\times10^{-6}\), 10 warmup steps, constant \\
    Weight decay & 0.01 \\
    Prompt / response limit & 4,096 / 64 tokens \\
    Rollout tensor parallelism & 4 \\
    EMA script update rate & 0.05 \\
    Distillation support & Student top 100 and a shared tail \\
    CPA & Layers 3 and 7, bias coefficient 0.05 \\
    Correction stabilizer / seed & \(10^{-8}\) / 20260913 \\
    \bottomrule
  \end{tabularx}
\end{minipage}
\end{center}

After accumulation and unscaling, shards contribute to a global dot product and retention squared norm. The shared correction precedes clipping and AdamW. The EMA teacher scores prefixes supplied by the rollout snapshot.

\section{Details of Dataset Generation}
\label{sec:data-generation}
\textbf{Annotations to questions.} Resolve references, apply a template, and map the geometric answer to its option index. Check visibility, unique reference, alignment, and distance from an ambiguous boundary.

\smallskip
\textbf{Labels to views.} The crop-to-JOINT mapping supplies the CPA prior. The teacher uses Dense crops if the gate fails. The student sees the overview.

\smallskip
\textbf{Counterfactual checks.} Recompute the index after permuting options. A horizontal mirror maps \(x\) to \(W-x\), reversing horizontal order. Recompute the label from the reflected boxes and retain the parent image family. The vertical order survives the mirror because \(y\) is unchanged.

%% file: supplement/sections/derivations.tex
\section{Supporting Derivations}
\label{sec:supp-derivations}
\begingroup
\setcounter{equation}{0}
\renewcommand{\theequation}{S\arabic{equation}}
\ifcsname theHequation\endcsname
  \renewcommand{\theHequation}{supp.\arabic{equation}}
\fi

\subsection{Coordinate order and attention odds}
\label{sec:supp-geometry-odds}

\textbf{Order under JOINT.}
Let \(A_J\) map an overview point \(u=(u_x,u_y)\) inside
the shared crop to JOINT. On axis \(d\in\{x,y\}\), write
\([A_J(u)]_d=a_d(u_d-o_d)+b_d\), where \([\cdot]_d\) selects
the axis, \(o_d\) is the crop origin in overview coordinates,
\(a_d>0\) is the resize factor, and \(b_d\) is the padding offset
in JOINT coordinates. For two overview box centers \(c_i,c_j\),
the shared offsets cancel, giving
\begin{equation}
 [A_J(c_i)-A_J(c_j)]_d = a_d(c_{i,d}-c_{j,d}).
 \label{eq:supp-joint-order}
\end{equation}

The positive factor preserves axis order and thus any strict
extremum. If each recovered center has error at most \(e_d\)
on that axis in overview units, the gap can change by at most
\(2e_d\). A true gap larger than \(2e_d\) in magnitude retains
its sign. At equality the pair can tie.

\smallskip
\textbf{Effect of CPA on one attention row.}
Fix an active head and a query token index \(a\) in crop \(i\).
For each visible key \(b\), let \(L_{ab}\) be its pre-bias logit
and \(A_{ab},\widetilde A_{ab}\) its softmax probabilities before
and after adding CPA to that fixed row. For a JOINT key \(p\),
\(G_i(p)\) is its correspondence prior mass (main Eq.~(5)).
The added logit is \(B_{ap}=\lambda_A\log(G_i(p)+\epsilon_A)\),
where \(\lambda_A>0\) is the bias scale and \(\epsilon_A>0\)
stabilizes the log. For visible JOINT keys \(p,p'\), the softmax
normalizers cancel, leaving the bias difference
\begin{equation}
 \log\frac{\widetilde A_{ap}}{\widetilde A_{ap'}}
 -\log\frac{A_{ap}}{A_{ap'}}
 =\lambda_A\log\frac{G_i(p)+\epsilon_A}{G_i(p')+\epsilon_A}.
 \label{eq:supp-cpa-odds}
\end{equation}

If \(p\) starts \(\Delta=L_{ap'}-L_{ap}>0\) logits behind
\(p'\), it overtakes \(p'\) when the prior ratio exceeds
\(\exp(\Delta/\lambda_A)\), with a tie at equality. This
comparison fixes the mask and pre-bias logits. Changes from earlier
CPA layers can also affect the logits of later layers.

\subsection{Information lost by tail aggregation}
\label{sec:supp-tail}

At a fixed prefix, let \(P,Q\) be the teacher and student
distributions over vocabulary \(\mathcal V\). Let \(S\) be the
student's fixed top-\(K_d\) set and \(T=\mathcal V\setminus S\).
Coarsening maps any distribution \(R\) to \(\widetilde R\) by
keeping its entries on \(S\) and replacing \(T\) with one bucket
of mass \(R(T)=\sum_{v\in T}R(v)\). For \(R(T)>0\),
let \(R_T(v)=R(v)/R(T)\) for \(v\in T\). Use mixture
\(M=\beta P+(1-\beta)Q\) with
teacher weight \(0<\beta<1\).

\smallskip
\textbf{KL partition.}
The terms on \(S\) coincide. On the tail, writing
\(P(v)/M(v)=[P(T)/M(T)]\allowbreak[P_T(v)/M_T(v)]\) splits the sum
into the bucket term and a conditional KL term~\cite{zhao2022dkd}, giving
\begin{equation}
 \begin{aligned}
 D_{\mathrm{KL}}(P\Vert M)
 &-D_{\mathrm{KL}}(\widetilde P\Vert\widetilde M)\\
 &=P(T)D_{\mathrm{KL}}(P_T\Vert M_T).
 \end{aligned}
 \label{eq:supp-tail-kl}
\end{equation}

\parfillskip=0pt plus .24\columnwidth
\lefthyphenmin=3
\righthyphenmin=3
\abovedisplayskip=7pt plus 1pt minus 2pt
\belowdisplayskip=7pt plus 1pt minus 2pt
\abovedisplayshortskip=2pt plus 1pt
\belowdisplayshortskip=4pt plus 1pt minus 1pt

\textbf{JSD gap.}
The full \(\operatorname{JSD}_{\beta}(P,Q)\) averages the KLs
from \(P\) and \(Q\) to \(M\) with weights \(\beta\) and
\(1-\beta\). Main Eq.~(9) uses the coarsened distributions.
Coarsening commutes with mixing. If \(M(T)>0\), conditioning on
\(T\) yields \(M_T=\beta_TP_T+(1-\beta_T)Q_T\), where
\(\beta_T=\beta P(T)/M(T)\). Applying Eq.~\eqref{eq:supp-tail-kl}
to both KLs and factoring out \(M(T)\) gives
\begin{equation}
 \begin{aligned}
 \operatorname{JSD}_{\beta}(P,Q)
 &-\operatorname{JSD}_{\beta}(\widetilde P,\widetilde Q)\\
 &=M(T)\operatorname{JSD}_{\beta_T}(P_T,Q_T).
 \end{aligned}
 \label{eq:supp-tail-js}
\end{equation}

With both tail masses positive, the gap is nonnegative and
vanishes exactly when \(P_T=Q_T\). Since
\(M_T\ge\beta_TP_T\) and \(M_T\ge(1-\beta_T)Q_T\), the
conditional KLs are bounded by \(-\log\beta_T\) and
\(-\log(1-\beta_T)\). Hence the gap is at most \(M(T)H_b(\beta_T)\),
where \(H_b(x)=-x\log x-(1-x)\log(1-x)\) is binary entropy. Terms with
zero mass are omitted. If either tail mass is zero, the gap is zero.
For \(P=(0.8,0.2,0)\), \(Q=(0.8,0,0.2)\), and \(\beta=1/2\),
retaining the first token yields zero coarsened JSD. The full JSD
reaches the upper bound \(0.2\log 2\) because the conditional tails
have disjoint support and the mixture assigns equal weight to both.

\smallskip
\textbf{What the tail mass controls.}
Since \(H_b(\beta_T)\le\log 2\), the total mass \(M(T)\) alone
yields the pointwise upper bound \(M(T)\log 2\), even when the two
conditional distributions are unknown. If their mixture puts at most
\(0.05\) of its probability in the tail, for example, the lost JSD
is at most \(0.05\log 2\) nats at that prefix. The tighter entropy
bound applies when one side supplies most of the tail mass.
For fixed \(S\) and \(P\), the coarsened loss is constant along
any redistribution of student tail probabilities that preserves
\(Q(T)\). The bucket registers the total probability assigned
to the tail, while Eq.~\eqref{eq:supp-tail-js} accounts for
disagreement over how that probability is allocated within it.

\subsection{When the sampled KL gradient is unbiased}
\label{sec:supp-retention}

At a given input and response prefix \(h\), let \(p(v)\) and
\(r(v)\) be the student and frozen reference probabilities of
token \(v\). Both are positive on the vocabulary. Let \(q(v)\)
be the actual rollout sampling probability at \(h\), including
any sampling filters. For student parameters \(\theta\), define
\(F_\theta(h)=D_{\mathrm{KL}}(p\Vert r)\),
\(f_v=\log[p(v)/r(v)]\), and \(s_v=\nabla_\theta\log p(v)\).
Given a scalar baseline \(b\) fixed at \(h\), differentiation with
\(r\) held fixed gives
\begin{equation}
 \begin{aligned}
 \nabla_\theta F_\theta(h)&=\sum_v p(v)(f_v+1)s_v\\
 &=\sum_v p(v)(f_v-b)s_v.
 \end{aligned}
 \label{eq:supp-retention-kl}
\end{equation}

The score identity
\(\sum_v p(v)s_v=\nabla_\theta\sum_v p(v)=0\) follows from
the normalization of \(p\). The two sums in
Eq.~\eqref{eq:supp-retention-kl} differ by
\((1+b)\sum_v p(v)s_v\), which vanishes at every prefix.
The baseline may therefore depend on \(h\) as long as it has
the same value for every vocabulary token at that
prefix~\cite{gu2024minillm}. A token-dependent baseline \(b(v)\)
would instead add the term \(-\sum_v p(v)b(v)s_v\), which
cancels only if this weighted score average is zero.

\smallskip
\textbf{Bias under the actual sampling law.}
With a detached coefficient, main Eq.~(12) contributes
\(\widehat g(v)=w(v)(f_v-b)s_v\) for a draw \(v\sim q\),
where \(w(v)\) is the effective importance weight.
Subtracting Eq.~\eqref{eq:supp-retention-kl} from its mean gives
\begin{equation}
 \begin{aligned}
 \mathbb E_{v\sim q}[\widehat g(v)]-\nabla_\theta F_\theta(h)
 ={}&\sum_v [q(v)w(v)-p(v)]\\[-2pt]
 &\hspace{7pt}\cdot(f_v-b)s_v.
 \end{aligned}
 \label{eq:supp-retention-bias}
\end{equation}

The bias is zero if \(q(v)>0\) wherever \(p(v)>0\) and
\(w(v)=p(v)/q(v)\). For excluded tokens use \(q(v)w(v)=0\).
Capping the ratio at \(c>0\) sets \(q(v)w(v)=\min\{p(v),cq(v)\}\).
Under truncation or clipping, a nonzero weighted score sum
\(\sum_v q(v)w(v)s_v\) makes the bias depend on the baseline.

\smallskip
\textbf{A bound from discarded probability.}
With truncation and optional upper clipping, each deficit
\(p(v)-q(v)w(v)\) is nonnegative when the ratio uses the actual
sampling law. If \(B\) bounds \(\|(f_v-b)s_v\|_2\) over tokens
with positive deficits, the triangle inequality in
Eq.~\eqref{eq:supp-retention-bias} bounds the gradient bias in
norm by \(B(1-\sum_v q(v)w(v))\). With an exact ratio on the
sampled support and no clipping, the factor in parentheses is
the student mass on excluded tokens. The bound depends both on
that mass and on the size of the omitted score contributions.
The rollout sampler determines which tokens are excluded in
this calculation. The top-\(K_d\) partition concerns the
separate distillation objective in main Eq.~(9).

\smallskip
\textbf{Variance-minimizing baseline.}
Under the unbiased conditions, the mean is independent of \(b\).
The covariance trace therefore differs from the second moment
\(\mathbb E_q[w(v)^2\|s_v\|_2^2(f_v-b)^2]\) by a constant in
\(b\), with \(\|\cdot\|_2\) the Euclidean norm. Setting the
derivative of that second moment to zero gives
\begin{equation}
 b^*=\frac{\mathbb E_q[w(v)^2\|s_v\|_2^2 f_v]}
                 {\mathbb E_q[w(v)^2\|s_v\|_2^2]}.
 \label{eq:supp-baseline}
\end{equation}

With finite moments and a positive denominator, the second
derivative in \(b\) is strictly positive, so
Eq.~\eqref{eq:supp-baseline} gives the unique minimizer.

\smallskip
\textbf{From a prefix to the rollout objective.}
The expectation conditions on the prefix before drawing its next
token. For the weighted batch sum, each weight must be fixed at
that point. In main Eq.~(12), \(a_{n,t}/Z_K\) uses the selector
\(a_{n,t}\) for token position \(t\) of sequence \(n\) and the
batch normalizer \(Z_K\). Either can affect the expectation if
it depends on the resulting token or completion. If prefix law
\(\rho_\theta(h)\) varies differentiably on locally fixed support,
the product rule gives
\begin{equation}
 \begin{aligned}
 \nabla_\theta\mathbb E_{h\sim\rho_\theta}[F_\theta(h)]
   ={}&\mathbb E_{h\sim\rho_\theta}[\nabla_\theta F_\theta(h)]\\
      &+\mathbb E_{h\sim\rho_\theta}
         [F_\theta(h)\nabla_\theta\log\rho_\theta(h)].
 \end{aligned}
 \label{eq:supp-prefix-law}
\end{equation}

The second term accounts for changes in prefix visitation.
Under the unbiased conditions above, main Eq.~(12) estimates
the first term with the rollout law fixed. Even exact next-token
ratios leave the visitation term when the prefix law itself is
differentiated.

\subsection{Guarantees of the stabilized correction}
\label{sec:supp-correction}

Let \(g_M\) be the distillation gradient from main Eq.~(10) and
\(g_K\) the sampled retention gradient from main Eq.~(12).
Write \(\gamma=\langle g_M,g_K\rangle\) and
\(\kappa=\|g_K\|_2^2\), using the Euclidean inner product and
norm. Along \(-g_M\) the sampled retention loss increases to
first order if \(\gamma<0\). Main Eq.~(13) sets
\(g_{\mathrm{final}}=g_M+t g_K\) with
\(t=[-\gamma]_+/(\kappa+\epsilon_p)\),
where \([x]_+=\max\{x,0\}\) and \(\epsilon_p>0\) is the
stabilizer. Thus
\begin{equation}
 \langle g_{\mathrm{final}},g_K\rangle
 =\gamma+\frac{\kappa[-\gamma]_+}{\kappa+\epsilon_p}.
 \label{eq:supp-residual-conflict}
\end{equation}

For \(\gamma<0\), the conflict magnitude shrinks to a fraction
\(\epsilon_p/(\kappa+\epsilon_p)\) of its initial value. The
inner product remains negative and the component orthogonal to
\(g_K\) is preserved. The remaining fraction depends on the
relative size of \(\epsilon_p\) and \(\kappa\). When the
retention gradient is zero, \(\gamma=0\) and the update is
\(g_M\).

\smallskip
\textbf{First-order progress on distillation.}
In the conflicting case \(\gamma<0\), we have \(\kappa>0\).
Decompose the original gradient as
\(g_M=g_\perp+(\gamma/\kappa)g_K\), where
\(\langle g_\perp,g_K\rangle=0\). Substituting the coefficient
\(t\) from main Eq.~(13) rescales the parallel term by
\(\epsilon_p/(\kappa+\epsilon_p)\). Orthogonality therefore yields
\begin{equation*}
 \langle g_M,g_{\mathrm{final}}\rangle
 =\|g_\perp\|_2^2
 +\frac{\epsilon_p}{\kappa+\epsilon_p}\frac{\gamma^2}{\kappa}.
\end{equation*}

Both terms are nonnegative, and the second is strictly positive in
the conflicting case for \(\epsilon_p>0\). For \(\gamma\ge0\),
the update stays at \(g_M\), giving \(\|g_M\|_2^2\). Thus
\(-g_{\mathrm{final}}\) is a first-order descent direction for
the local distillation loss whenever \(g_M\ne0\), using the
same teacher targets and selected support as the original
distillation gradient.

\smallskip
\textbf{Penalized projection.}
The corrected gradient uniquely minimizes a squared displacement
from \(g_M\) plus a penalty for negative alignment with \(g_K\),
given by the objective
\begin{equation}
 \frac{1}{2}\|g-g_M\|_2^2
   +\frac{[-\langle g,g_K\rangle]_+^2}{2\epsilon_p}.
 \label{eq:supp-soft-projection}
\end{equation}

If \(\gamma\ge0\), both terms vanish at \(g_M\). Otherwise,
an orthogonal displacement only increases the first term, so the
optimum has the form \(g=g_M+t g_K\). In the active region
\(\gamma+t\kappa<0\), stationarity gives
\(t+(\gamma+t\kappa)/\epsilon_p=0\). Its solution
\(t=-\gamma/(\kappa+\epsilon_p)\) satisfies that region's
inequality. Strict convexity gives uniqueness. In the limit
\(\epsilon_p\to0^+\), this is the one-sided projection of a
conflicting gradient pair used in PCGrad~\cite{yu2020pcgrad}.
For exactly antiparallel gradients the limiting projection is
zero because the orthogonal component vanishes. A positive
stabilizer retains a fraction of the original direction in that
case.

\smallskip
\textbf{Full retention gradient.}
Write \(F_{\mathcal H}(\theta)\) for the full-vocabulary loss
(main Eq.~(11)) on the same fixed prefixes \(\mathcal H\) and
with the same position weights. Its gradient is
\(g_K^*=\nabla_\theta F_{\mathcal H}(\theta)=g_K+e\), where
\(e\) measures the error incurred by using the sampled gradient
on that batch. Both gradients use the same loss normalization.

Combining Eq.~\eqref{eq:supp-residual-conflict} with
Cauchy--Schwarz bounds the directional derivative under the full
retention loss by
\begin{equation}
 -\langle g_K^*,g_{\mathrm{final}}\rangle
 \le \frac{\epsilon_p[-\gamma]_+}{\kappa+\epsilon_p}
      +\|e\|_2\|g_{\mathrm{final}}\|_2.
 \label{eq:supp-full-direction}
\end{equation}

The two terms account for residual conflict and estimation error.
In a batch where the sampled gradient is positively aligned with
the correction, keeping the signed term gives a sharper criterion.
Cauchy--Schwarz gives
\(\langle g_K^*,g_{\mathrm{final}}\rangle\ge
\langle g_K,g_{\mathrm{final}}\rangle
-\|e\|_2\|g_{\mathrm{final}}\|_2\).
The full retention loss therefore decreases to first order along
\(-g_{\mathrm{final}}\) if the sampled alignment exceeds the
error term. This criterion makes the role of the gradient estimate
explicit when the distillation and retention gradients agree.

\smallskip
\textbf{Actual optimizer step.}
Scalar clipping by a positive factor preserves the sign in
Eq.~\eqref{eq:supp-residual-conflict}. Coordinate-wise scaling
can reverse it. With \(g_K=(1,1)\), scaling \((2,-1)\) to
\((2,-3)\) changes the inner product from \(1\) to \(-1\).
Let \(\Delta\theta\) denote the actual AdamW displacement and
assume the gradient of \(F_{\mathcal H}\) is \(L\)-Lipschitz on
that segment. Integrating along the step gives
\begin{equation}
 \begin{aligned}
 F_{\mathcal H}(\theta+\Delta\theta)-F_{\mathcal H}(\theta)
 \le{}&\langle g_K^*,\Delta\theta\rangle\\
       &+\frac{L}{2}\|\Delta\theta\|_2^2.
 \end{aligned}
 \label{eq:supp-adamw-step}
\end{equation}

At step fraction \(s\), the gradient changes by at most
\(Ls\|\Delta\theta\|_2\). Its integrated contribution is the
quadratic remainder in Eq.~\eqref{eq:supp-adamw-step}.
A negative right-hand side certifies a finite-step decrease. For
a plain step \(\Delta\theta=-\eta g_{\mathrm{final}}\) with
positive full-gradient alignment and \(L>0\), a sufficient
step-size interval is
\begin{equation*}
 0<\eta<
 \frac{2\langle g_K^*,g_{\mathrm{final}}\rangle}
 {L\|g_{\mathrm{final}}\|_2^2}.
\end{equation*}

For AdamW, use the displacement after momentum, adaptive
preconditioning, and weight decay. Evaluate \(F_{\mathcal H}\)
before and after the step on the same prefixes and with the same
position weights and frozen reference in both evaluations.
For a nonzero displacement, Eq.~\eqref{eq:supp-adamw-step}
certifies decrease once the negative linear term per unit
squared step exceeds \(L/2\). This comparison uses the
actual direction and length of the AdamW update.
\endgroup

%% file: supplement/sections/cost.tex
\section{Cost}
Main Table 4 reports end-to-end times for 1,000 serial requests on one A100,
including loading, preprocessing and generation. Relative to Overview,
adding JOINT costs 36.72~s at 4B and 51.10~s at 9B. Adding CPA on top of
JOINT costs 0.94~s and 6.04~s, respectively.

\smallskip
The total increase over Overview is 37.66~s at 4B and 57.14~s at 9B.
JOINT accounts for about 98\% and 89\% of those increases, respectively.
At deployment, the student runs on Overview alone.

%% file: supplement/sections/cases.tex
\section{Qualitative Visualizations}
\label{sec:qualitative}
\noindent
\begin{minipage}[t]{.477\linewidth}
\setlength{\parindent}{1pc}
The white box in Overview marks the JOINT window. Crop colors match the boxes in JOINT. Center coordinates in each check are percentages of the JOINT width or height, measured from the left or top.
\end{minipage}\hfill
\begin{minipage}[t]{.477\linewidth}
\setlength{\parindent}{1pc}
The Overview text describes plausible errors. For archived cases, answer changes compare the JOINT and JOINT+CPA teachers. Gray padding marks a source edge with limited context.
\end{minipage}
\par\nopagebreak[4]\vspace{5bp}

\begin{CaseFigure}{blue}{Binding two similar fishing boats}
    {The teal cover identifies the left boat, and the green coil identifies the right boat.}
\CaseQuestion{Is the teal-covered boat left or right of the boat with the green coil?}
\CaseChoices{\Choice{A}{Left}\Choice{B}{Right}}
\CaseViews{\includegraphics[width=\linewidth,height=112bp,keepaspectratio]{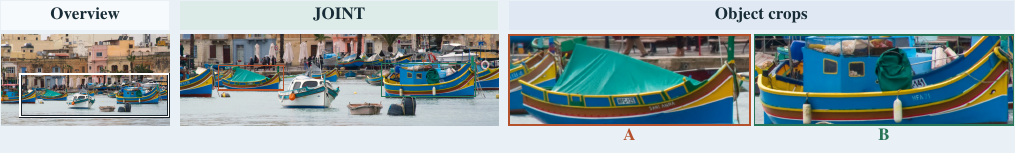}}
\CaseEvidence{Wrong answer}{B. Right}
  {Both boats share blue hulls and yellow trim, so color alone gives the same description for two distinct instances. If the teal cover is attributed to the boat with the green coil, the two identities swap while their locations stay fixed. This reverses the answer.}
  {Correct answer}{A. Left}
  {Crop A shows a flat sheet stretched over the hull. Crop B shows the raised loops of the coil. JOINT ties the sheet to the left-hand hull and the coil to the right-hand hull. The center comparison then yields Left.}
\CaseCheck{Horizontal order.}{\(x_A=22.3<x_B=78.1\).}{
  The 55.8-point separation is wide. Swapping the identities suffices to reverse the answer.

  With both hull identities fixed, the gap allows about 27.9 points of horizontal error per center before a tie is possible.
}
\end{CaseFigure}

\CaseSpacer

\begin{CaseFigure}{green}{Ordering three targets on a station façade}
    {The clock's center is above those of the station lettering and rail sign.}
\CaseQuestion{Which of the following has the smallest vertical image coordinate?}
\CaseChoices{\Choice{A}{Station-name lettering}\Choice{B}{Red rail-symbol sign}\Choice{C}{Round clock}}
\CaseViews{\includegraphics[width=\linewidth,height=102bp,keepaspectratio]{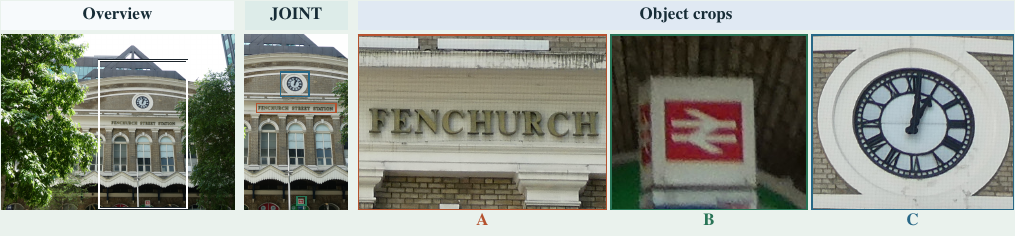}}
\CaseEvidence{Wrong answer}{A. Station-name lettering}
  {The name stretches across the façade, while the clock is a smaller feature near the top of the entrance. A search that prioritizes legible lettering can omit the dial. The station name then beats the red sign in a comparison of the remaining targets.}
  {Correct answer}{C. Round clock}
  {Crop C resolves the dial, A the block lettering, and B the rail symbol. JOINT places the dial above the name band and the red sign well below both. Since image \(y\) increases downward, the smaller center coordinate belongs to the clock.}
\CaseCheck{Vertical order.}{\(y_C=28.2<y_A=41.9<y_B=94.9\).}{
  The clock leads the lettering by 13.7 points of JOINT height.

  The nearest gap gives a radius of about 6.8 points per center for preserving the clock as the vertical minimum.
}
\end{CaseFigure}

\medskip
\noindent\textbf{Identifying targets in clutter}\par\nopagebreak[4]
First identify the intended instance, then compare its position with the other candidates. A shared crop and resize preserve horizontal order by Eq.~\eqref{eq:supp-joint-order}.
\par\nopagebreak[4]\vspace{4bp}

\begin{CaseFigure}{blue}{Distinguishing two similar drinks at a food counter}
    {The orange drink with the purple straw has the rightmost center.}
\CaseQuestion{Select the listed object with the largest horizontal image coordinate.}
\CaseChoices{\Choice{A}{the strawless orange drink}\Choice{B}{the drink with the purple straw}\Choice{C}{the salad bowl}}
\CaseViews{\includegraphics[width=\linewidth,height=112bp,keepaspectratio]{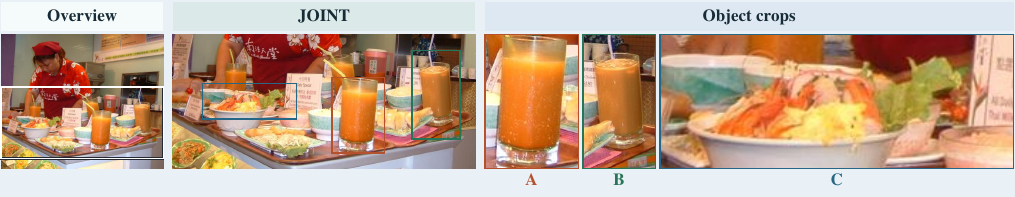}}
\CaseEvidence{Illustrative distractor}{A. the strawless orange drink}
  {The two orange drinks have nearly the same color. The larger, strawless glass is prominent in the foreground, so a quick scan can miss the glass with the purple straw. In that reduced comparison, the foreground glass lies to the right of the salad bowl.}
  {Geometric answer}{B. the drink with the purple straw}
  {Crop B shows the purple straw, crop A shows the strawless glass, and crop C shows the salad bowl. JOINT keeps the two glasses in one coordinate frame. Their box centers place the drink with the straw farthest right.}
\CaseCheck{Horizontal order.}{\(x_C=25.3<x_A=61.3<x_B=86.9\).}{
  The drink with the straw is 25.6 points of JOINT width right of the other glass.

  The closest box edges are separated by 9.0 points of JOINT width.
}
\end{CaseFigure}

\CaseSpacer

\begin{CaseFigure}{blue}{Matching similar furniture finishes}
    {The straight-edged table has the rightmost center, beyond the similarly finished counter.}
\CaseQuestion{Select the listed object with the largest horizontal image coordinate.}
\CaseChoices{\Choice{A}{the tables}\Choice{B}{the table}\Choice{C}{the counter}}
\CaseViews{\includegraphics[width=\linewidth,height=112bp,keepaspectratio]{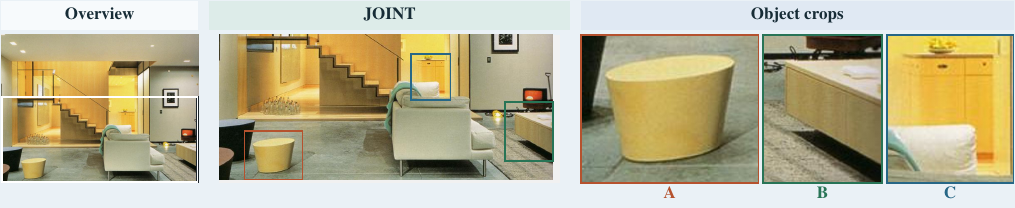}}
\CaseEvidence{Wrong answer}{C. the counter}
  {The table and counter near the right edge share a pale finish. Matching by finish can conflate the two. If the table is missed, the counter has the rightmost center among the remaining candidates, yielding C.}
  {Correct answer}{B. the table}
  {Shape distinguishes the similar finishes. A is round, B has a straight tabletop edge, and C shows vertical counter panels. JOINT places the table and counter in separate boxes, with B farther right.}
\CaseCheck{Horizontal order.}{\(x_A=16.0<x_C=62.0<x_B=90.8\).}{
  The table is 28.8 points farther right than the counter, its closest rival for the maximum.

  The measured gap gives a radius of about 14.39 points of horizontal error per center for the table and the counter.

  Below that radius, B remains the maximum when all three furniture candidates are matched to their annotated boxes.
}
\end{CaseFigure}

\CaseSpacer
\begin{CaseFigure}{blue}{A small target beside a vehicle}
    {The woman in red has the rightmost center, beyond the steering wheel and the group of people.}
\CaseQuestion{Select the listed object with the largest horizontal image coordinate.}
\CaseChoices{\Choice{A}{the people}\Choice{B}{the woman}\Choice{C}{the steering wheel}}
\CaseViews{\includegraphics[width=\linewidth,height=112bp,keepaspectratio]{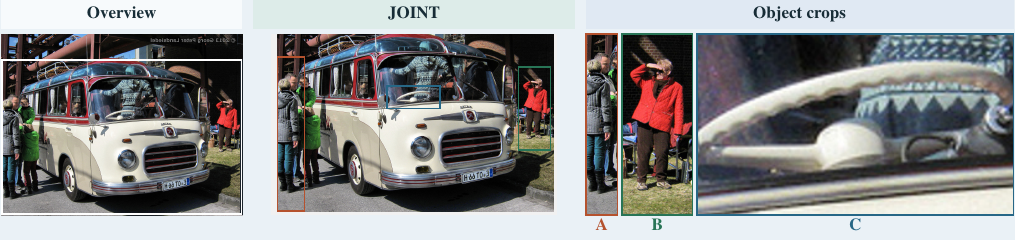}}
\CaseEvidence{Wrong answer}{C. the steering wheel}
  {The vehicle dominates the Overview and draws the search to its windshield. If the woman on the right is missed, the wheel wins the comparison against the group of people.}
  {Correct answer}{B. the woman}
  {Crop B identifies the red coat, while C resolves the wheel through the windshield. JOINT places both in the same frame as the group. The coat belongs to the rightmost box, so the ranking selects the woman.}
\CaseCheck{Horizontal order.}{\(x_A=6.9<x_C=49.8<x_B=92.2\).}{
  The woman's center leads the wheel by 42.4 points, measured before rounding the centers.

  The wheel leads the group by 42.9 points when B is absent, so the reduced set has a clear maximum.

  Restoring the pedestrian changes the answer while leaving the centers of the other two candidates fixed.
}
\end{CaseFigure}

\CaseSpacer

\begin{CaseFigure}{blue}{Finding the leftmost target on a shopping street}
    {The pedestrian's center lies to the left of both the book display and the street light.}
\CaseQuestion{Select the listed object with the smallest horizontal image coordinate.}
\CaseChoices{\Choice{A}{the street light}\Choice{B}{the man}\Choice{C}{the books}}
\CaseViews{\includegraphics[width=\linewidth,height=112bp,keepaspectratio]{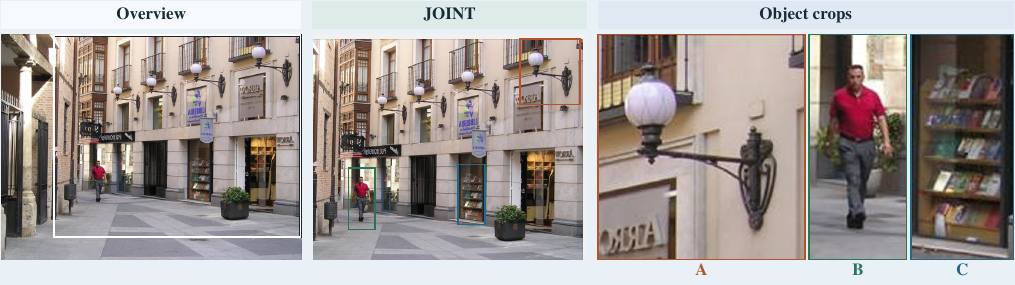}}
\CaseEvidence{Wrong answer}{C. the books}
  {The book display and lamp belong to the shopfront. A search focused on the façade can miss the pedestrian. The books sit left of the lamp and win that incomplete comparison.}
  {Correct answer}{B. the man}
  {The red shirt in B, book spines in C and curved bracket in A anchor the three boxes. The pedestrian is farther back but has the smallest horizontal center. The image-axis comparison depends on the box centers, regardless of depth.}
\CaseCheck{Horizontal order.}{\(x_B=17.7<x_C=58.0<x_A=86.8\).}{
  The man's center lies 40.3 points to the left of the book display.

  In the two-candidate comparison, the books lead the lamp by 28.7 points even though both boxes lie on the shopfront.

  Adding the pedestrian introduces a center to the left of that entire interval and therefore changes the minimum.
}
\end{CaseFigure}

\CaseSpacer
\begin{CaseFigure}{green}{Selecting a horizontal extreme among four targets}
    {The small café sign has the rightmost center, beyond the bicycle rack.}
\CaseQuestion{Which target has the largest \(x\)?\enskip
  \Choice{A}{Blue hatchback}\Choice{B}{Double-headed lamp}\Choice{C}{Bicycle rack}\Choice{D}{Café sign}}
\CaseViews{\includegraphics[width=\linewidth,height=112bp,keepaspectratio]{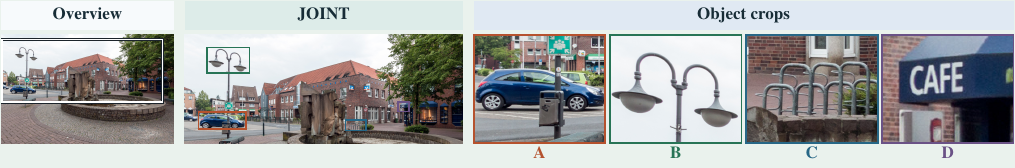}}
\CaseEvidence{Wrong answer}{C. Bicycle rack}
  {The fountain draws attention to the center, and the bicycle rack sits beyond it. The café sign is a small patch under the far-right tree. A search that ends at the fountain leaves out that region.

  The rack would then win a comparison among the car, lamp and rack. That answer is consistent with the visible subset. The search has stopped before all four named objects have been located.}
  {Correct answer}{D. Café sign}
  {Crop D isolates the white café lettering, and crop C shows repeated metal loops. These local cues distinguish the two rightmost candidates, even though the sign takes up little of the scene.

  In JOINT the car and lamp are close on the left, followed by the rack and the sign. Their close ordering on the left has no bearing on the maximum. Comparing the two rightmost candidates selects D.}
\CaseCheck{Horizontal order.}{\(x_A=13.6<x_B=15.7<x_C=61.5<x_D=78.6\).}{
  The sign clears the rack by 17.1 points. The car and lamp are separated by only 2.1 points.
}
\end{CaseFigure}

\CaseSpacer

\begin{CaseFigure}{green}{Locating the leftmost object in a kitchen}
    {The glass cake stand's center lies to the left of the woven basket and lidded jar.}
\CaseQuestion{Select the listed object with the smallest horizontal image coordinate.}
\CaseChoices{\Choice{A}{the lidded jar}\Choice{B}{the glass cake stand}\Choice{C}{the woven basket}}
\CaseViews{\includegraphics[width=\linewidth,height=112bp,keepaspectratio]{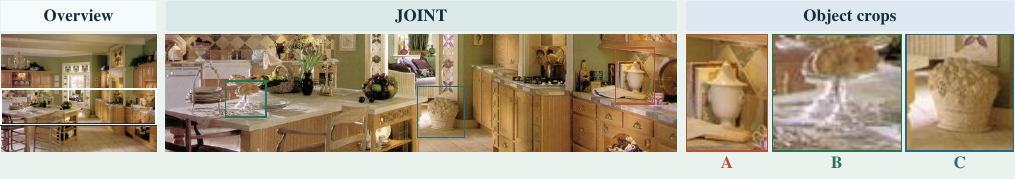}}
\CaseEvidence{Illustrative distractor}{C. the woven basket}
  {The transparent cake stand blends into the reflections and glassware on the island. The woven basket has a distinct silhouette near the doorway and is easier to pick out in the full scene.

  If the stand is missed, the basket lies left of the jar and becomes the leftmost remaining candidate.}
  {Geometric answer}{B. the glass cake stand}
  {Crop B shows the clear pedestal and the pastry it supports. The basket in C sits near the doorway. The lidded jar in A is on the right counter.

  In JOINT, the cake stand sits to the left of both objects. Its box center is the leftmost of the three.}
\CaseCheck{Horizontal order.}{\(x_B=15.7<x_C=53.9<x_A=91.8\).}{
  The cake stand is 38.2 points of JOINT width left of the basket, its nearest rival.

  The closest box edges are separated by 29.5 points of JOINT width.
}
\end{CaseFigure}